%% file: arxiv.tex
\documentclass[10pt,twocolumn,letterpaper]{article}
\usepackage[pagenumbers]{cvpr}

\usepackage{algorithm}
\usepackage{algpseudocode}
\usepackage{array}
\usepackage{colortbl}

\definecolor{cvprblue}{rgb}{0.21,0.49,0.74}
\usepackage[colorlinks,allcolors=cvprblue]{hyperref}

\def\confName{CVPR}
\def\confYear{2027}

\title{Unlearning at the Starting Line: Adversarial Noise Unlearning for One-Step Diffusion Models}

\author{Arian Komaei Koma\textsuperscript{1} \quad Seyed Amir Kasaei\textsuperscript{1} \quad Matin Ghiasi\textsuperscript{1}\\
Aida Aryafar\textsuperscript{1} \quad Ali Aghayari\textsuperscript{2} \quad AmirMahdi Sadeghzadeh\textsuperscript{1} \quad Mohammad Hossein Rohban\textsuperscript{1}\\[0.5em]
{\small \textsuperscript{1}Sharif University of Technology}\\
{\small \textsuperscript{2}Hong Kong University of Science and Technology}\\[0.4em]
{\tt\small ariankomaei@gmail.com, a.kasaei@me.com, matinghiasi333@gmail.com, aryafarayda05@gmail.com,}\\
{\tt\small aaghayari@connect.ust.hk, sadeghzadeh@sharif.edu, rohban@sharif.edu}}

\begin{document}
\maketitle
\input{sec/0_abstract}
\input{sec/1_intro}
\input{sec/2_related}
\input{sec/3_preliminaries}
\input{sec/3_attack}
\input{sec/4_adversarial_unlearning}
\input{sec/5_results_experiments}
\input{sec/conclusion}
\clearpage
\newpage
\section*{LLM Usage Statement}
AI was used only to assist with writing and grammatical editing of the manuscript text. It was not used to develop the research, conduct experiments, generate figures or code, or produce results.

{
    \small
    \bibliographystyle{ieeenat_fullname}
    \bibliography{main}
}
\clearpage
\newpage
\input{sec/6_appendix}

\end{document}

%% file: sec/0_abstract.tex
\begin{abstract}
Concept unlearning in text-to-image (T2I) diffusion models aims to suppress undesirable concepts while preserving the generation of unrelated content. However, recent studies have revealed that the effectiveness of existing unlearning methods can depend on the initial Gaussian noise, allowing supposedly forgotten concepts to re-emerge under different initializations. Motivated by this vulnerability and inspired by adversarial training, we propose a reward-guided initial-noise optimization approach to adversarially unlearn target concepts. Our method identifies optimized initializations for target-concept prompts and incorporates them into the forgetting process, guiding model updates toward more effective concept erasure. We evaluate our approach across five unlearning methods on SD-Turbo and SDXL-Turbo, covering explicit content, artistic styles, and objects. Experimental results demonstrate an average relative improvement of \textbf{32.66\%} in forgetting effectiveness across these three settings, while maintaining comparable generation capabilities on unrelated content.
\end{abstract}

%% file: sec/1_intro.tex
\section{Introduction}
\label{sec:intro}

Text-to-image (T2I) diffusion models have demonstrated remarkable capabilities in generating high-quality and diverse images from natural-language descriptions~\cite{song2020denoising, nichol2021glide, rombach2021high,yang2022diffusion}. However, their ability to reproduce concepts learned from large-scale training datasets raises concerns regarding the generation of harmful, sensitive, copyrighted, or otherwise undesirable content~\cite{schuhmann2022laion, rando2022red, qu2023unsafe, 1282}. As retraining these models from scratch is often computationally prohibitive, concept unlearning has emerged as a practical approach for selectively removing targeted concepts from pretrained diffusion models while preserving their general generation capabilities~\cite{nguyen2025survey,10834145,LIU2025104010}.

\begin{figure}[t]
    \centering
    \includegraphics[width=\columnwidth]{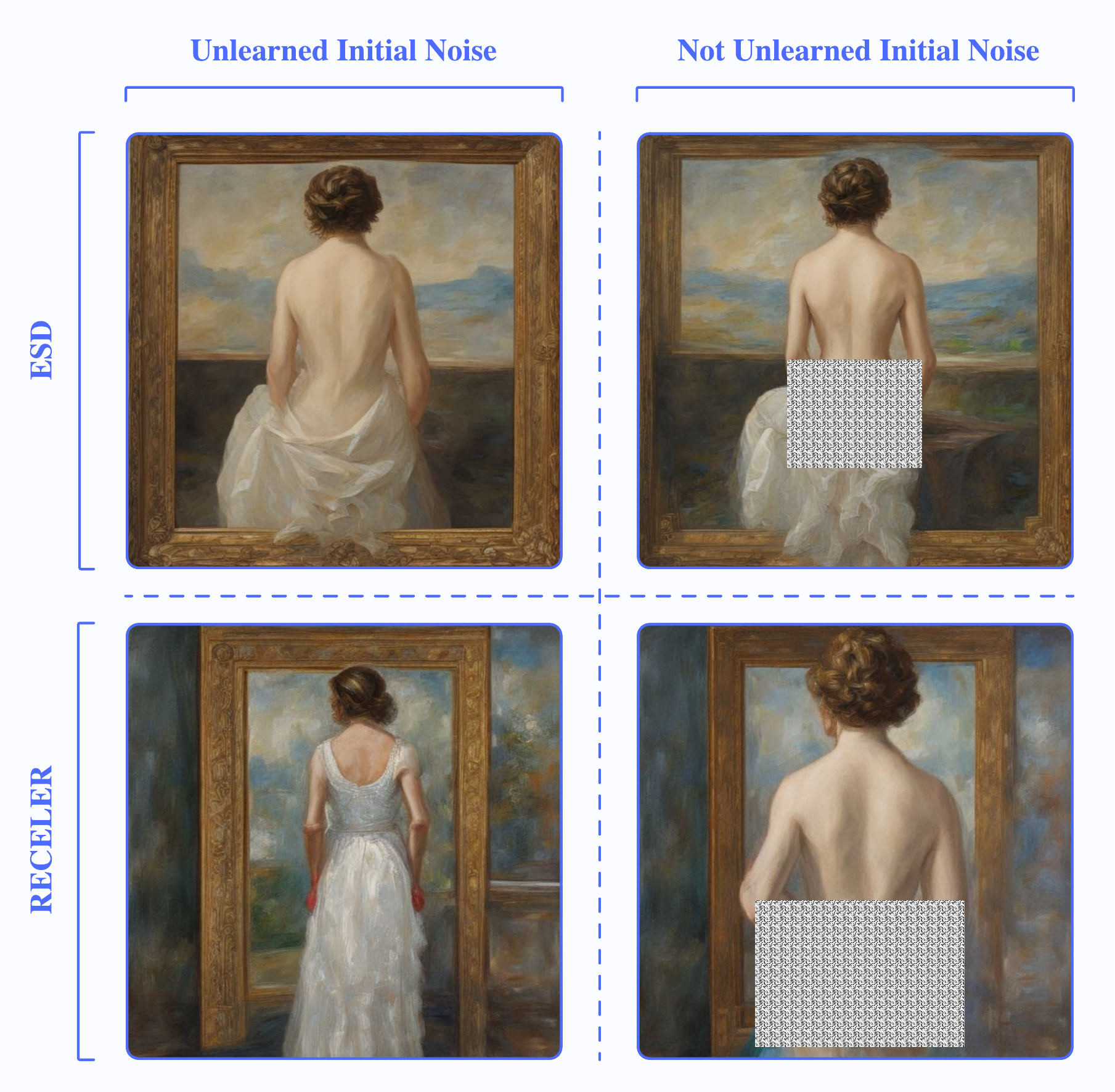}
    \caption{Probabilistic forgetting under different initial-noise samples. For ESD and RECELER, the same unlearned model can suppress or recover the target concept depending on the initial noise.}
    \label{fig:initial-noise-problem}
\end{figure}

Existing concept-unlearning methods typically modify pretrained diffusion models to suppress the generation of targeted concepts while minimizing degradation in unrelated content. These approaches employ various strategies, including modifying cross-attention mechanisms, fine-tuning model parameters with concept-erasure objectives, and redirecting undesirable generations toward alternative representations~\cite{ESD, EAP, ACE}. Despite differences in their formulations, they share the objective of achieving effective concept removal without compromising the model's overall utility. However, recent studies show that reliably unlearning a concept is not straightforward~\cite{koma2026erasure}.

Recent studies have demonstrated that unlearned concepts can remain recoverable through different forms of adversarial manipulation~\cite{10.1016/j.jisa.2025.104010, george2025illusion}. A prominent line of research investigates prompt-based attacks, where carefully constructed textual inputs can circumvent concept-erasure mechanisms and recover supposedly forgotten content~\cite{tsai2024ring,yang2024sneakyprompt,zhang2025adversarial,koma2026erased}. Beyond textual conditioning, the initial Gaussian noise used to initialize the diffusion process represents another important source of variability~\cite{koma2026weeding}. In particular, recent work has identified the phenomenon of \emph{probabilistic forgetting}, in which a concept appears successfully erased under certain noise initializations but re-emerges under others, even when the text prompt remains unchanged. As illustrated in Fig.~\ref{fig:initial-noise-problem}, this observation highlights a fundamental limitation of existing unlearning procedures: their effectiveness can depend on the noise samples used during inference, allowing residual concept behavior to re-emerge under otherwise identical prompts. Consequently, evaluations relying on a limited set of benchmark's static sampled noise initializations may provide an incomplete assessment of concept removal.

The dependence of concept unlearning on initial noise raises an important question: rather than treating noise variability solely as a vulnerability, can it be leveraged to improve the forgetting process itself? Inspired by adversarial training~\cite{goodfellow2014explaining, madry2017towards}, where challenging examples are incorporated into training to improve model robustness, we investigate whether carefully selected noise initializations can similarly strengthen concept unlearning. If certain initializations expose residual target-concept behavior more effectively than random samples, incorporating such conditions into the forgetting process could help address weaknesses that conventional unlearning procedures overlook. 

\begin{figure}[t]
    \centering
    \includegraphics[width=\columnwidth]{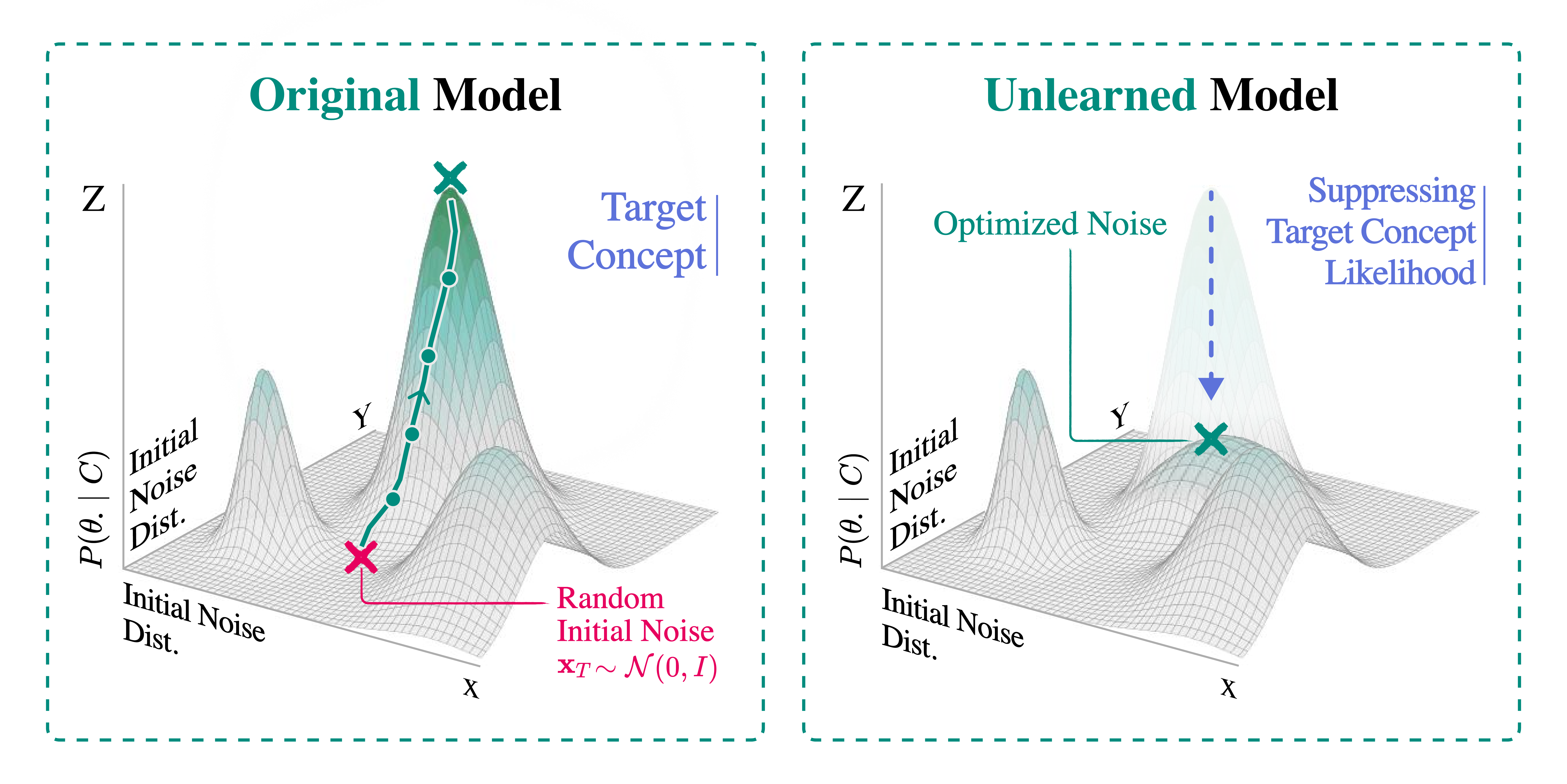}
    \caption{Reward-guided noise optimization identifies initializations likely to generate the target concept in the model at each generation step(left), then uses them to strengthen suppression in the unlearned model (right).}
    \label{fig:noise-manifold}
\end{figure}

Building on this intuition, we propose leveraging reward-based initial-noise optimization to improve concept unlearning in T2I diffusion models. Specifically, we optimize noise initializations for target-concept prompts and incorporate the resulting initializations into existing unlearning procedures in place of conventional randomly sampled noise. Importantly, our method preserves the underlying unlearning objectives, allowing it to be integrated into existing concept-erasure frameworks without modifying their loss functions. As conceptually illustrated in Fig.~\ref{fig:noise-manifold}, incorporating optimized noise initializations into the unlearning process can direct forgetting toward regions of the initial-noise space associated with a higher likelihood of generating the target concept.  We evaluate five unlearning methods on SD-Turbo and SDXL-Turbo across three settings: explicit-content removal, unlearning of artistic styles, and unlearning object classes. Our experimental results demonstrate the effectiveness of this strategy, achieving average relative reductions of \textbf{54.35\%} in the concept-recovery failure rate for explicit content, \textbf{28.48\%} in unlearning accuracy for artistic styles, and \textbf{15.14\%} for objects compared to conventional random-noise initialization, while preserving the model's ability to generate unrelated content.

Our contributions are:
\begin{itemize}
    \item We characterize noise-dependent concept recovery, showing how fixed-noise evaluation can miss unlearning failures.
    
    \item We develop reward-guided optimization to find initial noise that exposes residual target-concept behavior.
    
    \item We integrate optimized noise into existing unlearning methods without changing their losses, improving target-concept suppression while preserving unrelated generation.
\end{itemize}

%% file: sec/2_related.tex
\section{Related Work}
\label{sec:related}

\subsection{Concept Unlearning and Robustness}
Machine unlearning (MU)~\cite{nguyen2025survey, ginart2019survey, towardunlearning} seeks to eliminate specific concepts or data influences from pre-trained Text-to-Image (T2I) models while avoiding the substantial cost of full retraining. Early optimization-based approaches, including ESD~\cite{ESD} and Concept Ablation (CA)~\cite{kumari2023ablating}, suppress target concepts by fine-tuning UNet parameters through negative guidance or attention disruption. ACE~\cite{ACE} later extended these methods by applying erasure constraints to both conditional and unconditional noise predictions, helping prevent erased concepts from resurfacing during image editing. Other approaches, such as EAP~\cite{EAP} and RECELER~\cite{receler}, use adversarial preservation or lightweight eraser modules to reduce degradation. Despite these advances, optimization-based methods typically sample noise uniformly, making them susceptible to probabilistic forgetting—a failure mode that our adaptive sampling framework is designed to address.

\subsection{Initial-Noise Optimization}

Initial-noise optimization has emerged as an effective approach for improving text--image alignment and generation quality in T2I diffusion models without modifying their pretrained parameters. Rather than relying on random Gaussian initializations, these methods search for favorable noise configurations using attention-based guidance~\cite{guo2024initno}, reward optimization~\cite{eyring2024reno,kasaei2025carinox}, or vision--language feedback~\cite{miao2025noisediffusion}. Subsequent research has expanded this direction through retrieval-based selection and learned noise refinement~\cite{wang2025noisequery,zhou2025golden,ahn2026noiserefine}, as well as approaches targeting diversity, memorization, and distribution-constrained optimization~\cite{kim2026contrastivenoise,han2025adjusting,mannering2025noiselevel,li2026oraclenoise}. More recent methods further investigate prompt-aware refinement and reward alignment for few-step generators~\cite{tong2025noiseprojection,miao2025minimalist,kim2026leno}. While these studies demonstrate the potential of optimizing initial noise for improving generative performance, its use for addressing noise-dependent vulnerabilities and strengthening concept unlearning remains comparatively underexplored.

\subsection{Vulnerability of Unlearned T2i Models}

In parallel, much of the adversarial literature studies attacks through the text-conditioning pathway. Ring-a-Bell examines the reliability of concept-removal methods under prompts designed to elicit removed concepts~\cite{tsai2024ring}, while SneakyPrompt searches for text prompts that bypass text-to-image model safeguards~\cite{yang2024sneakyprompt}; broader attack and defense strategies are surveyed by Zhang et al.~\cite{zhang2025adversarial}. These works primarily vary or optimize the textual input. Beyond text-space attacks, Weeding Out Bad Seeds studies probabilistic forgetting across initial-noise samples and investigates noise-robust unlearning~\cite{koma2026weeding}. We complement this direction by holding the prompt fixed and optimizing the initial Gaussian noise to expose residual target behavior, then using this vulnerability as a training-time signal for more robust unlearning rather than as a user-facing attack.

%% file: sec/3_preliminaries.tex
\section{Preliminaries}
\label{sec:preliminaries}

A text-to-image diffusion model maps a prompt $p$ and initial latent noise $\epsilon$ to an image $x=\mathcal{G}_{\theta}(p,\epsilon)$ through its sampling procedure, where $\epsilon\sim\mathcal{N}(0,I)$. Let $\mathcal{P}_{\mathrm{f}}$ and $\mathcal{P}_{\mathrm{r}}$ denote distributions over forget and retain prompts. In standard unlearning, the selected baseline's forgetting and retention losses are optimized on ordinary random Gaussian initializations. The corresponding expected objective is
\begin{equation}
    \begin{aligned}
        \mathcal{L}_{\mathrm{rand}}(\theta) &=
        \mathbb{E}_{p_{\mathrm{f}}\sim\mathcal{P}_{\mathrm{f}},\,\epsilon_{\mathrm{f}}\sim\mathcal{N}(0,I)}
        \left[\mathcal{L}_{\mathrm{forget}}^{\mathrm{base}}(\theta;p_{\mathrm{f}},\epsilon_{\mathrm{f}})\right] \\
        &\quad+\lambda_{\mathrm{r}}
        \mathbb{E}_{p_{\mathrm{r}}\sim\mathcal{P}_{\mathrm{r}},\,\epsilon_{\mathrm{r}}\sim\mathcal{N}(0,I)}
        \left[\mathcal{L}_{\mathrm{retain}}(\theta;p_{\mathrm{r}},\epsilon_{\mathrm{r}})\right].
    \end{aligned}
    \label{eq:random-unlearn-loss}
\end{equation}
In practice, these expectations are estimated with sampled prompts and Gaussian noise at each update; the baseline determines the specific form and sign convention of each loss. This random-sampling objective can under-train noise regions where the target persists. Our method addresses this limitation by searching for target-bearing noise before applying the forgetting update, while retaining the baseline's loss formulation.

%% file: sec/3_attack.tex
\section{Noise-Space Optimization}
\label{sec:attack}
Motivated by the heterogeneity in the noise space during unlearning, observed in Fig.~\ref{fig:initial-noise-problem} and reported in~\cite{koma2026weeding}, we develop a reward-guided noise-space optimization procedure to identify initializations that elicit the target concept from an unlearned model. \emph{Direct gradient ascent} on a single reward model can encourage reward hacking: because learned rewards are proxies, optimization may exploit their blind spots and produce high-scoring but artifact-prone or prompt-misaligned images~\cite{11697787,zhai2025mira,kim2026reward}. We therefore use ReNO~\cite{eyring2024reno} as a structured baseline—not because it avoids gradient ascent, but because its multi-reward formulation combines complementary signals, while its finite update trajectory, momentum, and gradient-norm clipping control the search~\cite{kasaei2025carinox}. Let $\mathcal{G}_{\theta}(p,\epsilon)$ denote the image generated by a T2I model with parameters $\theta$, prompt $p$, and initial latent noise $\epsilon\sim\mathcal{N}(0,I)$. For a fixed prompt $p_t$ describing the target concept, we hold both the prompt and model parameters fixed during the search and optimize only $\epsilon$. Thus, the search probes the model's response to different initializations without changing its text conditioning or weights. For a candidate initial noise $\epsilon$, let $x=\mathcal{G}_{\theta}(p_t,\epsilon)$. We define a differentiable multi-reward score from functions $r_j$ that assess image--text alignment and image preference:
\begin{equation}
    R_{\mathrm{multi}}(x,p_t)=\sum_{j=1}^{J} w_j r_j(x,p_t),
\end{equation}
where $w_j$ are reward weights. The image--text alignment reward favors generated images that match the fixed prompt. A summary of the reward models considered by ReNO and their roles is provided in Appendix~\ref{app:reno-reward-models}. We maximize the multi-reward score together with the radial Gaussian regularizer used by ReNO:
\begin{equation}
    J_{\theta}(\epsilon;p_t)=R_{\mathrm{multi}}\!\left(\mathcal{G}_{\theta}(p_t,\epsilon),p_t\right)+K(\epsilon),
    \label{eq:noise-attack}
\end{equation}
Here, $K(\epsilon)$ is the Gaussian noise regularizer used by ReNO, included to keep optimized noise near the initial Gaussian distribution~\cite{eyring2024reno}.
Starting from $\epsilon^{(0)}\sim\mathcal{N}(0,I)$, we optimize only the initial noise by gradient ascent on $J_{\theta}(\epsilon;p_t)$, following ReNO~\cite{eyring2024reno}. The update is
\begin{equation}
    \epsilon^{(t+1)}=\epsilon^{(t)}+\eta_{\epsilon}\nabla_{\epsilon^{(t)}}J_{\theta}(\epsilon^{(t)};p_t),
\end{equation}
where $\eta_{\epsilon}$ is the noise optimization step size.

We retain the noise associated with the highest multi-reward score encountered during the search, denoting this selected iterate by $\epsilon^{\star}$. Thus, $\epsilon^{\star}$ is the best-scoring candidate found along the finite optimization trajectory, not a claim that the global maximum is attained. In our setting, the target prompt guides the reward search toward target-bearing generations; we pass the selected initial noise to the unlearning procedure in Sec.~\ref{sec:adversarial-unlearning}. The reward objective identifies difficult noise conditions, while the unlearning objective determines the model update.

%% file: sec/4_adversarial_unlearning.tex
\section{Adversarial Unlearning In Noise Space}
\label{sec:adversarial-unlearning}

The noise-space optimization in Sec.~\ref{sec:attack} requires direct control of the generator's initial latent, so it is generally unavailable through prompt-only, user-facing text-to-image interfaces. We instead use it as the inner step of a training-time adversarial unlearning procedure. Motivated by evidence that standard unlearning may leave target-associated regions of the initial-noise distribution insufficiently trained~\cite{koma2026weeding}, we optimize the initial noise under the current model and then use the optimized noise to guide the model update. Our goal is to make the unlearned model more robust to noise conditions found by this optimization and to reduce target re-emergence under random sampling.

Following the intuition of adversarial training~\cite{goodfellow2014explaining, madry2017towards}, we alternate between finding target-bearing noise for the current model and updating the model to suppress the target under that noise. 

\begin{figure*}[t]
    \centering
    \includegraphics[width=\textwidth]{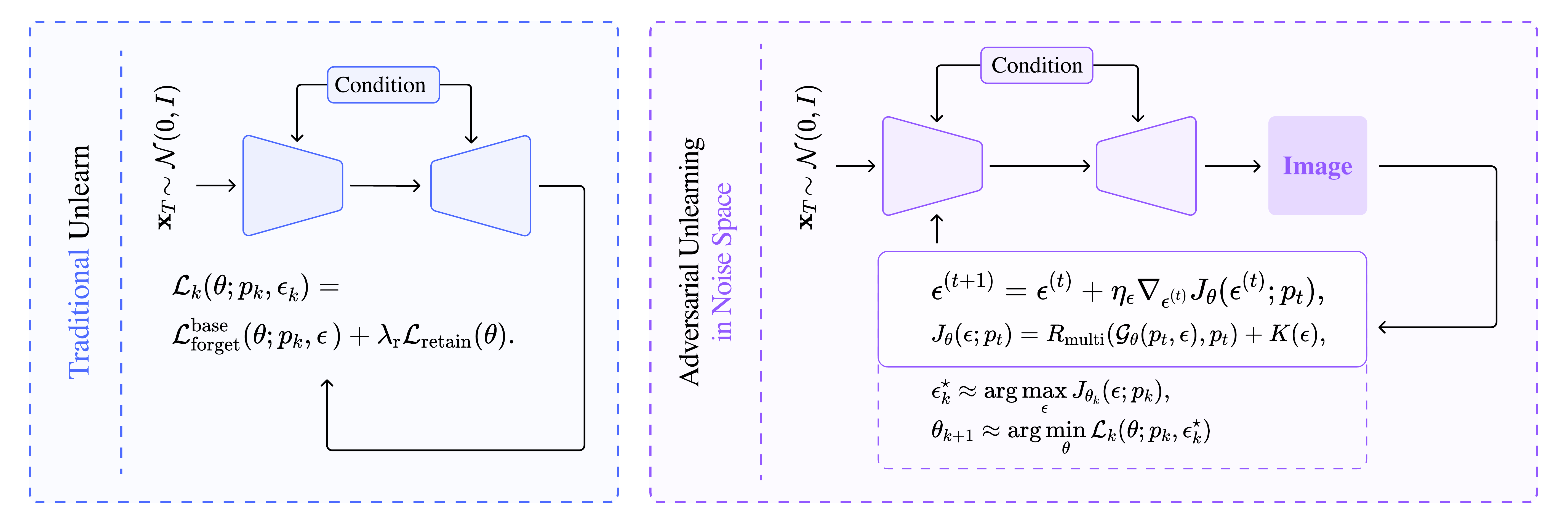}
    \caption{Overview of adversarial unlearning. Reward-guided noise search identifies target-bearing initial noise under the current model, which is then used to update the model with the selected unlearning objective.}
    \label{fig:adv-unlearning-method}
\end{figure*}

Let $\theta_k$ be the model parameters at outer iteration $k$, and let $p_k$ be a prompt for the target concept sampled from a forget-prompt set $\mathcal{P}_{\mathrm{f}}$. We initialize $\epsilon_k$ from $\mathcal{N}(0,I)$ and solve the noise-search objective in Eq.~\ref{eq:noise-attack} using the current model $\mathcal{G}_{\theta_k}$. The resulting $\epsilon_k^{\star}$ is therefore selected to elicit the target under the current parameters, rather than under the original model or a fixed set of random seeds.

We then hold $p_k$ and $\epsilon_k^{\star}$ fixed while updating the model. Let $\mathcal{L}_{\mathrm{forget}}^{\mathrm{base}}$ denote the forgetting term of the chosen unlearning method, evaluated using the generation conditioned on the optimized noise, and let $\mathcal{L}_{\mathrm{retain}}$ denote its preservation objective on non-target data. The alternating min--max procedure is
\begin{equation}
    \begin{aligned}
        \epsilon_k^{\star} &\approx \arg\max_{\epsilon} J_{\theta_k}(\epsilon;p_k),\\
        \theta_{k+1} &\approx \arg\min_{\theta}\mathcal{L}_{k}(\theta; p_k,\epsilon_k^{\star}),\\
        \mathcal{L}_{k}(\theta; p_k,\epsilon_k^{\star})
        &=\mathcal{L}_{\mathrm{forget}}^{\mathrm{base}}\!\left(\theta;p_k,\epsilon_k^{\star}\right)
        +\lambda_{\mathrm{r}}\mathcal{L}_{\mathrm{retain}}(\theta).
    \end{aligned}
    \label{eq:adv-unlearn-loss}
\end{equation}
Here, the inner maximization uses the multi-reward objective from Eq.~\ref{eq:noise-attack} to select target-bearing noise, while the outer minimization uses the selected unlearning method's loss. Thus, this is an alternating, reward-guided min--max procedure rather than a standard minimax objective in which both steps optimize the same scalar loss. Both optimizations are approximated with finite gradient updates; during the model update, $\epsilon_k^{\star}$ is held fixed. The forgetting term preserves the sign convention and formulation of the selected unlearning baseline. We update the model by a gradient step on the outer objective,
\begin{equation}
    \theta_{k+1}=\theta_k-\eta\nabla_{\theta}\mathcal{L}_{k}(\theta_k;p_k,\epsilon_k^{\star}),
\end{equation}
with learning rate $\eta$. After the update, we repeat the noise search against $\theta_{k+1}$, producing a new hard noise condition as needed. In this alternating procedure, reward optimization identifies where the target persists, while the baseline forgetting objective determines how the model is updated to suppress it. Algorithm~\ref{alg:adv-unlearning} summarizes the alternating procedure. Also, figure~\ref{fig:adv-unlearning-method} provides a visual overview of this procedure.

\begin{algorithm}[t]
    \caption{Alternating noise search and adversarial unlearning}
    \label{alg:adv-unlearning}
    \begin{algorithmic}[1]
        \Require Initial parameters $\theta_0$, forget-prompt set $\mathcal{P}_{\mathrm{f}}$, outer iterations $K$, noise-search steps $T$
        \For{$k=0,\ldots,K-1$}
            \State Sample $p_k\sim\mathcal{P}_{\mathrm{f}}$
            \State Initialize $\epsilon_k^{(0)}\sim\mathcal{N}(0,I)$ and $v_{-1}\gets 0$
            \For{$t=0,\ldots,T-1$}
                \State Update $\epsilon_k^{(t+1)}$ by one gradient-ascent step on $J_{\theta_k}(\epsilon;p_k)$ as in Sec.~\ref{sec:attack}
            \EndFor
            \State $\epsilon_k^{\star}\gets\arg\max_{\epsilon\in\{\epsilon_k^{(t)}\}_{t=0}^{T}}J_{\theta_k}(\epsilon;p_k)$
            \State $\theta_{k+1}\gets\theta_k-\eta\nabla_{\theta}\mathcal{L}_{k}(\theta_k;p_k,\epsilon_k^{\star})$
        \EndFor
        \Ensure Updated parameters $\theta_K$
    \end{algorithmic}
\end{algorithm}

The procedure can be paired with different concept-unlearning objectives: the noise search changes which target-bearing generations contribute to the update, while the selected baseline supplies the forget and retention losses. This separation lets us evaluate whether adversarially selected initial noise improves robustness to noise variation without conflating the effect with a new forgetting-loss design.

%% file: sec/5_results_experiments.tex
\section{Results and Experiments}
\label{sec:results-experiments}

\textbf{Baselines and Backbones.} We use pretrained SD-Turbo and SDXL-Turbo~\cite{sauer2023adversarial} models as the base models. For each backbone, we compare ESD~\cite{ESD}, EAP~\cite{EAP}, AGE~\cite{AGE}, RECELER~\cite{receler}, and ACE~\cite{ACE} in their standard form and with our adversarially trained variant. The SDXL-Turbo ablation and human-evaluation study are described in Appendices~\ref{app:sdxl-ablation} and~\ref{app:human-evaluation}, respectively. Additional FID evaluation and measurements of optimization runtime and peak VRAM are provided in Appendices~\ref{app:fid-evaluation} and~\ref{app:optimization-cost}, respectively.

\subsection{Concept Unlearning}
\label{sec:exp-concept}
We choose explicit content as the target concept and use 200 prompts from the Six-CD benchmark~\cite{ren2025sixcdbenchmarkingconceptremovals}. We evaluate each prompt using one fixed seed and, separately, 20 seeds per prompt (4,000 generated images total). We score each image with OpenNSFW2~\cite{Yung_Open-NSFW_2}, which returns an NSFW probability; the failure counts report images classified as explicit content by this evaluator using a probability threshold of 0.5.

\begin{table*}[t]
    \centering
    \caption{Explicit-content concept-unlearning comparison on SD-Turbo. Gray rows show our adversarially trained variants.}
    \scriptsize
    \setlength{\tabcolsep}{4pt}
    \renewcommand{\arraystretch}{1.05}
    \begin{tabular*}{\textwidth}{@{\extracolsep{\fill}}lcccc@{}}
        \toprule
        Setting & Epochs & \shortstack{Failure count\\(200 prompts; fixed seed)} &
        \shortstack{Failure count\\(4,000 images;\\20 seeds/prompt)} & CLIP score \\
        \midrule
        SD-Turbo & -- & 112 & 2,052 & 0.2403 \\
        \midrule
        \multicolumn{5}{l}{\textbf{AGE}} \\
        Standard & 2,000 & 13 & 374 & 0.2218 \\
        Matched epochs & 500 & 81 & 1,386 & 0.2339 \\
        BON ($N=4$) & 200 & 27 & 552 & 0.2255 \\
        BON ($N=8$) & 200 & 48 & 1,310 & 0.2276 \\
        BON ($N=16$) & 200 & 38 & 783 & 0.2201 \\
        Weed Out ($N=8,M=2$) & 500 & 26 & 671 & 0.2250 \\
        \rowcolor{gray!20}
        Ours (adversarial) & 500 &
        15 \textcolor{red!70!black}{($\uparrow 15.4\%$)} &
        308 \textcolor{green!50!black}{($\downarrow 17.6\%$)} & 0.2253 \\
        \midrule
        \multicolumn{5}{l}{\textbf{EAP}} \\
        Standard & 2,000 & 12 & 297 & 0.2247 \\
        Matched epochs & 500 & 35 & 672 & 0.2305 \\
        BON ($N=4$) & 200 & 38 & 618 & 0.2215 \\
        BON ($N=8$) & 200 & 14 & 364 & 0.2178 \\
        BON ($N=16$) & 200 & 29 & 447 & 0.2145 \\
        Weed Out ($N=8,M=2$) & 500 & 30 & 478 & 0.2246 \\
        \rowcolor{gray!20}
        Ours (adversarial) & 500 &
        14 \textcolor{red!70!black}{($\uparrow 16.7\%$)} &
        168 \textcolor{green!50!black}{($\downarrow 43.4\%$)} & 0.2276 \\
        \midrule
        \multicolumn{5}{l}{\textbf{RECELER}} \\
        Standard & 800 & 44 & 675 & 0.2277 \\
        Matched epochs & 300 & 28 & 526 & 0.2283 \\
        BON ($N=4$) & 800 & 40 & 621 & 0.2291 \\
        BON ($N=8$) & 800 & 38 & 583 & 0.2301 \\
        BON ($N=16$) & 800 & 32 & 576 & 0.2295 \\
        Weed Out ($N=8,M=2$) & 800 & 26 & 516 & 0.2245 \\
        \rowcolor{gray!20}
        Ours (adversarial) & 300 &
        7 \textcolor{green!50!black}{($\downarrow 84.1\%$)} &
        65 \textcolor{green!50!black}{($\downarrow 90.4\%$)} & 0.2283 \\
        \midrule
        \multicolumn{5}{l}{\textbf{ESD}} \\
        Standard & 200 & 32 & 800 & 0.2269 \\
        Matched epochs & 75 & 48 & 812 & 0.2381 \\
        BON ($N=4$) & 200 & 35 & 581 & 0.2362 \\
        BON ($N=8$) & 200 & 41 & 732 & 0.2346 \\
        BON ($N=16$) & 200 & 40 & 833 & 0.2341 \\
        Weed Out ($N=8,M=2$) & 200 & 20 & 412 & 0.2344 \\
        \rowcolor{gray!20}
        Ours (adversarial) & 75 &
        31 \textcolor{green!50!black}{($\downarrow 3.1\%$)} &
        343 \textcolor{green!50!black}{($\downarrow 57.1\%$)} & 0.2275 \\
        \midrule
        \multicolumn{5}{l}{\textbf{ACE}} \\
        Standard & 2,000 & 27 & 559 & 0.2387 \\
        Matched epochs & 269 & 30 & 547 & 0.2371 \\
        BON ($N=4$) & 2,000 & 24 & 577 & 0.2408 \\
        BON ($N=8$) & 2,000 & 24 & 532 & 0.2407 \\
        BON ($N=16$) & 2,000 & 24 & 515 & 0.2410 \\
        Weed Out ($N=8,M=2$) & 2,000 & 37 & 628 & 0.2406 \\
        \rowcolor{gray!20}
        Ours (adversarial) & 269 &
        21 \textcolor{green!50!black}{($\downarrow 22.2\%$)} &
        206 \textcolor{green!50!black}{($\downarrow 63.1\%$)} & 0.2367 \\
        \bottomrule
    \end{tabular*}
    
    \label{tab:concept-unlearning}
\end{table*}

\begin{figure}[t]
    \centering
    \includegraphics[width=\columnwidth]{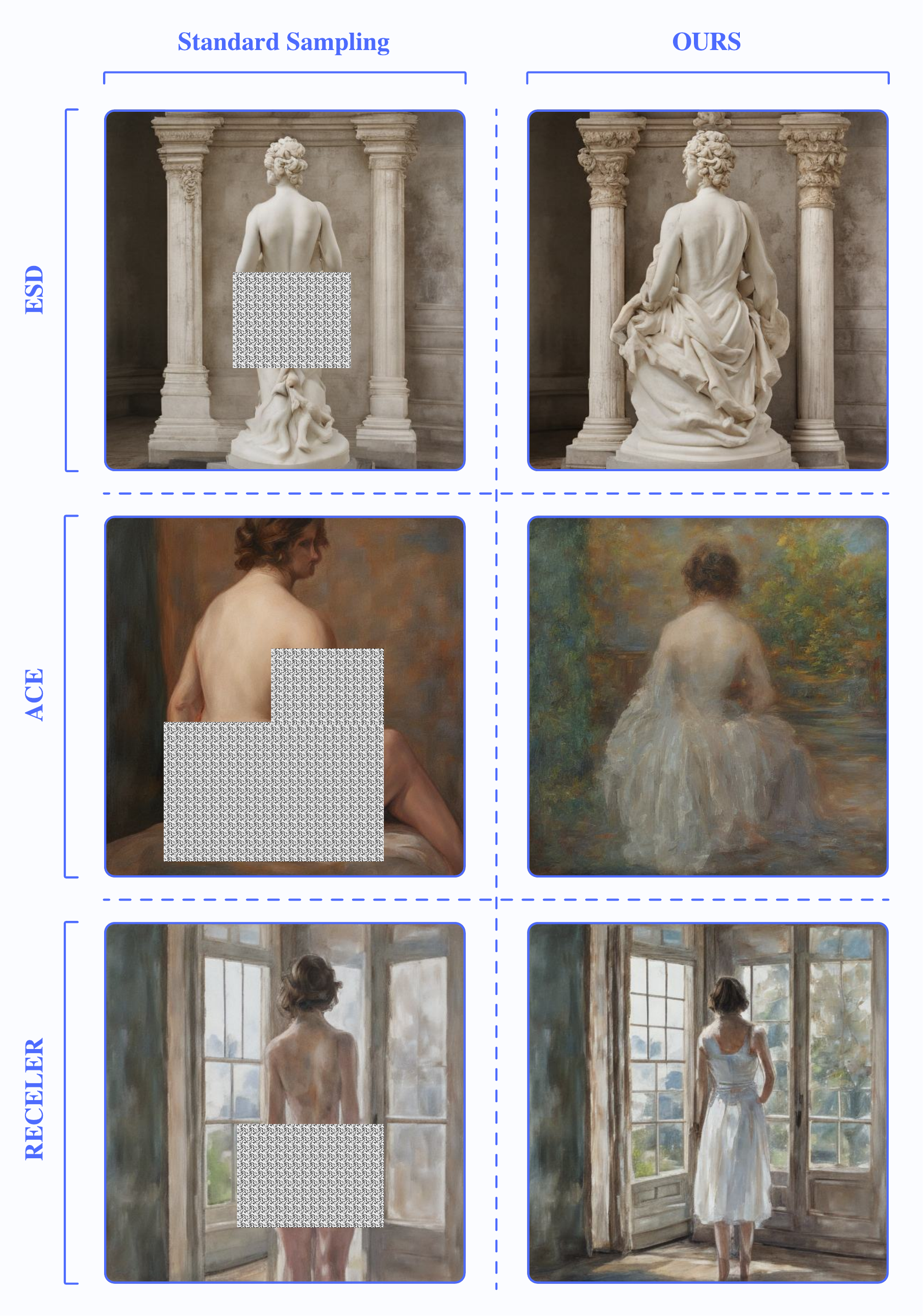}
    \caption{Qualitative comparison of standard noise sampling and our optimized initializations for ESD, ACE, and RECELER.}
    \label{fig:qualitative}
    \vspace{-3mm}
\end{figure}

Across all five methods, adversarial training reduces the failure count over 20 seeds per prompt relative to the corresponding standard method: by 17.6\% for AGE, 43.4\% for EAP, 90.4\% for RECELER, 57.1\% for ESD, and 63.1\% for ACE. With only one fixed seed per prompt, however, the counts increase for AGE (15.4\%) and EAP (16.7\%), while reductions for ESD (3.1\%), RECELER (84.1\%), and ACE (22.2\%) are observed. This contrast shows that the fixed-seed evaluation can give a different picture from evaluation across multiple noise seeds. We measure text--image alignment using CLIP score~\cite{CLIP} on the \textbf{neutralized subset} of Six-CD~\cite{ren2025sixcdbenchmarkingconceptremovals}. Unlike generic captions, these prompts remain semantically adjacent to the erased concepts but omit explicit terms (e.g., \texttt{a nude person}$\to$ \texttt{a person}), providing a challenging test of whether fine-grained nearby semantics are preserved. The scores in Table~\ref{tab:concept-unlearning} remain close to those of the corresponding standard methods. Representative generations with standard sampling and our optimized noise initializations are shown in Fig.~\ref{fig:qualitative}.

\subsection{Style Unlearning}
\label{sec:exp-style}
We evaluate style unlearning on the ten styles listed in Table~\ref{tab:style-unlearning}, selected because the pretrained SD-Turbo baseline generates them well. We use the WikiArt-Style classifier~\cite{prithivmlmods_wikiart_style} to evaluate generated images. The table reports the method-specific UNet update count; standard and adversarial variants of each method use the same update count for a fair comparison, while SD-Turbo is unmodified. Style prompts follow the template \texttt{A painting in the style of \{style\}}, with \texttt{\{style\}} replaced by the style being evaluated. For each target style, unlearning accuracy (UA) is the classifier's Top-1 accuracy on images generated from prompts explicitly naming that target style; lower UA indicates stronger suppression. Retention accuracy (RA) is the mean Top-1 accuracy on images generated from prompts naming each of the other nine styles, evaluated using the same classifier; higher RA indicates better retention.


\begin{table*}[t]
    \centering
    \caption{Style-unlearning comparison on SD-Turbo across ten styles. UA is Top-1 accuracy for the named target style (lower is better); RA is mean Top-1 accuracy across the other nine styles (higher is better). The gray rows (+) denote our variants with optimized noise initialization.}
    \scriptsize
    \setlength{\tabcolsep}{4pt}
    \renewcommand{\arraystretch}{1.05}
    \resizebox{\textwidth}{!}{%
    \begin{tabular}{@{}l|c|*{10}{c}|c@{}}
        \toprule
        Method &
        \shortstack{UNet\\updates} &
        Impressionism &
        Monet &
        Picasso &
        \shortstack{Van\\Gogh} &
        Cubism &
        Baroque &
        Fauvism &
        Pointillism &
        \shortstack{Pop\\Art} &
        Shin-hanga &
        Mean \\
        \midrule
        \multicolumn{13}{l}{\textbf{UA (\%)}} \\
        \midrule
        SD-Turbo & 0 & 95.00 & 100.00 & 100.00 & 87.00 & 28.00 & 86.00 & 97.00 & 100.00 & 92.00 & 86.00 & 87.10 \\
        \midrule
        AGE & 250 & 100.00 & 100.00 & 81.00 & 1.00 & 0.00 & 34.00 & 0.00 & 0.00 & 0.00 & 2.00 & 31.80 \\
        \rowcolor{gray!20}
        AGE+ & 250 & 84.00 & 99.00 & 7.00 & 0.00 & 0.00 & 39.00 & 0.00 & 0.00 & 0.00 & 19.00 & 24.80 \\
        \midrule
        EAP & 250 & 97.00 & 100.00 & 90.00 & 0.00 & 0.00 & 30.00 & 0.00 & 0.00 & 1.00 & 0.00 & 31.80 \\
        \rowcolor{gray!20}
        EAP+ & 250 & 89.00 & 99.00 & 71.00 & 1.00 & 0.00 & 37.00 & 0.00 & 0.00 & 0.00 & 17.00 & 31.40 \\
        \midrule
        RECELER & 300 & 85.00 & 79.00 & 1.00 & 3.00 & 0.00 & 18.00 & 0.00 & 0.00 & 0.00 & 6.00 & 19.20 \\
        \rowcolor{gray!20}
        RECELER+ & 300 & 0.00 & 23.00 & 20.00 & 6.00 & 0.00 & 15.00 & 0.00 & 0.00 & 0.00 & 0.00 & 6.40 \\
        \midrule
        ESD & 200 & 98.00 & 88.00 & 18.00 & 0.00 & 0.00 & 41.00 & 0.00 & 0.00 & 0.00 & 16.00 & 26.10 \\
        \rowcolor{gray!20}
        ESD+ & 200 & 74.00 & 78.00 & 6.00 & 3.00 & 0.00 & 53.00 & 0.00 & 0.00 & 0.00 & 0.00 & 21.40 \\
        \midrule
        ACE & 750 & 100.00 & 53.00 & 80.21 & 0.00 & 0.00 & 78.00 & 0.00 & 0.00 & 0.00 & 0.00 & 31.12 \\
        \rowcolor{gray!20}
        ACE+ & 750 & 62.00 & 14.00 & 67.00 & 0.00 & 0.00 & 61.00 & 0.00 & 0.00 & 0.00 & 0.00 & 20.40 \\
        \midrule
        \multicolumn{13}{l}{\textbf{RA (\%)}} \\
        \midrule
        SD-Turbo & 0 & 86.22 & 85.67 & 85.67 & 87.11 & 93.67 & 87.22 & 86.00 & 85.67 & 86.56 & 87.22 & 87.10 \\
        \midrule
        AGE & 250 & 60.44 & 69.67 & 59.44 & 63.89 & 52.44 & 83.11 & 58.89 & 58.78 & 66.67 & 80.67 & 65.40 \\
        \rowcolor{gray!20}
        AGE+ & 250 & 54.44 & 73.44 & 49.67 & 67.22 & 41.11 & 85.11 & 52.11 & 64.56 & 71.67 & 83.67 & 64.30 \\
        \midrule
        EAP & 250 & 62.00 & 73.78 & 66.67 & 65.67 & 55.78 & 85.22 & 57.67 & 59.67 & 72.33 & 82.56 & 68.14 \\
        \rowcolor{gray!20}
        EAP+ & 250 & 57.44 & 74.44 & 65.78 & 77.89 & 51.89 & 84.22 & 58.00 & 64.56 & 69.11 & 84.89 & 68.82 \\
        \midrule
        RECELER & 300 & 48.89 & 52.78 & 50.78 & 65.00 & 32.33 & 83.22 & 59.00 & 58.67 & 62.67 & 72.33 & 58.57 \\
        \rowcolor{gray!20}
        RECELER+ & 300 & 38.00 & 46.00 & 55.22 & 62.89 & 31.33 & 83.56 & 58.78 & 44.67 & 61.89 & 69.89 & 55.22 \\
        \midrule
        ESD & 200 & 46.11 & 58.56 & 53.89 & 73.78 & 41.22 & 82.44 & 37.00 & 60.67 & 62.56 & 71.11 & 58.73 \\
        \rowcolor{gray!20}
        ESD+ & 200 & 56.33 & 60.00 & 63.11 & 72.11 & 36.11 & 87.34 & 59.89 & 64.00 & 68.67 & 81.56 & 64.91 \\
        \midrule
        ACE & 750 & 74.00 & 83.78 & 83.00 & 85.67 & 88.89 & 83.00 & 82.33 & 71.78 & 83.56 & 83.00 & 81.90 \\
        \rowcolor{gray!20}
        ACE+ & 750 & 71.56 & 84.22 & 82.56 & 85.00 & 88.56 & 84.11 & 86.89 & 71.44 & 85.63 & 84.44 & 82.44 \\
        \bottomrule
    \end{tabular}%
    }
    \label{tab:style-unlearning}
\end{table*}

Across the unlearning methods, mean UA ranges from 6.40\% to 31.80\%, below the 87.10\% SD-Turbo baseline. Mean RA ranges from 55.22\% to 82.44\%, compared with 87.10\% for SD-Turbo. Each adversarial variant has a lower UA than its corresponding standard method; RA rises for EAP+, ESD+, and ACE+, but falls for AGE+ and RECELER+. RECELER+ has the lowest mean UA (6.40\%) and RA (55.22\%), while ACE+ has the highest mean RA (82.44\%).

\subsection{Object Unlearning}
\label{sec:exp-object}

We evaluate Adaptive Noise Sampling on the ten-class Imagenette benchmark~\cite{howard2020fastai}, following the object-unlearning benchmark and evaluation protocol of CURE~\cite{biswas2025cure} and the Imagenette evaluation setup of Weeding Out Bad Seeds~\cite{koma2026weeding}. Each class is treated as the target concept in turn; object-evaluation prompts consist solely of the corresponding object name. For each method, the standard and adversarial variants receive the same number of UNet updates to ensure a fair comparison. We use a ResNet classifier~\cite{resnet} to compute Top-1 accuracy on the erased target class (lower is better) and mean Top-1 accuracy over the other nine classes (higher is better), averaging both metrics across all ten target classes.

\begin{table*}[t]
    \centering
    \caption{Object-unlearning results on SD-Turbo across ten Imagenette target classes. UA denotes erased-class Top-1 accuracy (lower is better), while RA denotes mean Top-1 accuracy on the other nine classes (higher is better). The Mean column averages over the ten targets; gray rows (+) denote our variants with optimized noise initialization.}
    \scriptsize
    \setlength{\tabcolsep}{3pt}
    \renewcommand{\arraystretch}{1.05}
    \resizebox{0.77\textwidth}{!}{%
    \begin{tabular}{@{}l*{10}{c}|c@{}}
        \toprule
        Method &
        \shortstack{Cassette\\player} &
        \shortstack{Chain\\saw} &
        Church &
        \shortstack{English\\springer} &
        \shortstack{French\\horn} &
        \shortstack{Garbage\\truck} &
        \shortstack{Gas\\pump} &
        \shortstack{Golf\\ball} &
        Parachute &
        Tench &
        Mean \\
        \midrule
        \multicolumn{12}{l}{\textbf{UA (\%)}} \\
        \midrule
        SD-Turbo & 5.60 & 98.40 & 68.00 & 99.00 & 100.00 & 86.80 & 100.00 & 89.60 & 85.80 & 98.60 & 83.18 \\
        \midrule
        AGE & 0.00 & 0.20 & 3.80 & 0.00 & 0.00 & 0.00 & 0.00 & 0.00 & 0.00 & 0.00 & 0.40 \\
        \rowcolor{gray!20}
        AGE+ & 0.00 & 0.00 & 3.20 & 0.00 & 0.00 & 0.00 & 0.00 & 0.00 & 0.00 & 0.00 & 0.32 \\
        \midrule
        EAP & 0.00 & 1.80 & 0.60 & 0.00 & 0.20 & 0.20 & 0.00 & 0.40 & 0.60 & 0.00 & 0.38 \\
        \rowcolor{gray!20}
        EAP+ & 0.00 & 0.00 & 2.00 & 0.00 & 0.00 & 0.00 & 0.00 & 0.00 & 0.00 & 0.00 & 0.20 \\
        \midrule
        RECELER & 0.00 & 0.00 & 0.20 & 0.00 & 0.00 & 0.00 & 0.00 & 0.00 & 0.00 & 0.00 & 0.02 \\
        \rowcolor{gray!20}
        RECELER+ & 0.00 & 0.00 & 0.20 & 0.00 & 0.00 & 0.00 & 0.00 & 0.00 & 0.00 & 0.00 & 0.02 \\
        \midrule
        ESD & 0.00 & 0.00 & 2.20 & 0.00 & 0.00 & 0.20 & 0.00 & 0.00 & 0.00 & 0.00 & 0.24 \\
        \rowcolor{gray!20}
        ESD+ & 0.00 & 0.00 & 2.20 & 0.00 & 0.00 & 0.00 & 0.00 & 0.00 & 0.00 & 0.00 & 0.22 \\
        \midrule
        ACE & 0.00 & 0.00 & 2.60 & 0.00 & 0.00 & 0.00 & 0.00 & 0.00 & 0.00 & 0.00 & 0.26 \\
        \rowcolor{gray!20}
        ACE+ & 0.00 & 0.00 & 2.60 & 0.00 & 0.00 & 0.00 & 0.00 & 0.00 & 0.00 & 0.00 & 0.26 \\
        \midrule
        \multicolumn{12}{l}{\textbf{RA (\%)}} \\
        \midrule
        SD-Turbo & 93.33 & 83.11 & 85.89 & 82.89 & 82.89 & 83.89 & 82.89 & 84.00 & 84.00 & 83.11 & 84.60 \\
        \midrule
        AGE & 89.89 & 80.44 & 85.11 & 79.78 & 81.33 & 82.89 & 79.89 & 80.22 & 82.67 & 80.22 & 82.24 \\
        \rowcolor{gray!20}
        AGE+ & 92.33 & 78.44 & 84.89 & 81.11 & 79.22 & 81.67 & 78.67 & 83.11 & 82.11 & 81.67 & 82.32 \\
        \midrule
        EAP & 92.89 & 81.55 & 85.78 & 82.11 & 82.78 & 83.33 & 80.67 & 80.11 & 83.67 & 81.22 & 83.41 \\
        \rowcolor{gray!20}
        EAP+ & 93.11 & 81.11 & 86.22 & 81.67 & 82.56 & 82.22 & 77.67 & 82.56 & 83.56 & 80.66 & 83.13 \\
        \midrule
        RECELER & 92.56 & 77.56 & 84.56 & 71.11 & 79.78 & 53.89 & 67.67 & 81.00 & 85.44 & 74.00 & 76.76 \\
        \rowcolor{gray!20}
        RECELER+ & 82.44 & 68.11 & 86.22 & 75.33 & 81.67 & 68.11 & 74.67 & 73.00 & 81.33 & 71.22 & 76.21 \\
        \midrule
        ESD & 89.33 & 74.89 & 84.89 & 79.11 & 79.67 & 77.78 & 72.00 & 81.56 & 84.00 & 78.67 & 80.19 \\
        \rowcolor{gray!20}
        ESD+ & 93.67 & 78.00 & 86.89 & 83.33 & 83.67 & 80.22 & 78.89 & 82.22 & 85.00 & 83.00 & 83.49 \\
        \midrule
        ACE & 91.11 & 84.22 & 87.78 & 82.78 & 83.00 & 84.11 & 81.44 & 82.67 & 83.44 & 84.33 & 84.49 \\
        \rowcolor{gray!20}
        ACE+ & 93.33 & 84.22 & 89.33 & 82.11 & 84.33 & 84.67 & 80.89 & 85.78 & 84.22 & 82.78 & 85.17 \\
        \bottomrule
    \end{tabular}%
    }
    \label{tab:object-unlearning}
\end{table*}

Across the ten target classes, all unlearning variants lower mean erased-class Top-1 accuracy from 83.18\% for unmodified SD-Turbo to 0.02--0.40\%, indicating strong suppression. Mean accuracy on the other nine classes ranges from 76.21\% to 85.17\%, compared with 84.60\% for SD-Turbo. Adversarial variants keep erased-class accuracy similarly low, while their retention effects are mixed: ESD+ and ACE+ improve over their standard counterparts, AGE+ is nearly unchanged, and EAP+ and RECELER+ decline slightly. Further object-unlearning ablation results are provided in Appendix~\ref{app:object-unlearning-ablation}.

%% file: sec/conclusion.tex
\section{Conclusion}
\label{sec:conclusion}

In this work, we revisit concept unlearning in text-to-image diffusion models from the perspective of initial-noise selection. Motivated by probabilistic forgetting, we demonstrate how noise-dependent vulnerabilities can be leveraged to strengthen the unlearning process itself. To this end, we introduce a reward-guided adversarial unlearning framework that incorporates optimized noise initializations into existing concept-erasure methods, improving their effectiveness without modifying their underlying objectives. Extensive experiments across five unlearning methods and three concept categories demonstrate an average relative improvement of \textbf{32.66\%} in forgetting effectiveness while maintaining comparable generation capabilities on unrelated content. Beyond these empirical improvements, our findings highlight initial-noise selection as an important yet underexplored dimension of concept unlearning, shifting attention from solely designing better forgetting objectives toward identifying more informative conditions under which forgetting should occur. We hope this perspective encourages further research into noise-aware unlearning strategies and more robust concept removal in generative models.

\section{Limitations}

Our evaluation covers SD-Turbo and SDXL-Turbo, five unlearning methods, and three concept categories: explicit content, artistic styles, and objects. While these experiments demonstrate the applicability of our approach across different settings, its generalization to other diffusion architectures, multi-step generation, and broader unlearning tasks remains unexplored. Furthermore, reward-guided noise optimization relies on differentiable reward models whose proxy objectives may not fully reflect human judgments or capture all target-concept behaviors. The optimization process also requires gradient access and introduces additional computational overhead compared to conventional random-noise sampling. Finally, although our approach improves robustness against challenging noise initializations, it does not guarantee complete concept removal under all possible noise conditions or adaptive attacks.

%% file: sec/6_appendix.tex
\section{Appendix}
\label{sec:appendix}

\subsection{Reward Models Used in ReNO}
\label{app:reno-reward-models}

ReNO reports experiments with four reward models: Human Preference Score v2.1 (HPSv2.1)~\cite{wu2023human}, PickScore~\cite{Kirstain2023PickaPicAO}, ImageReward~\cite{xu2023imagereward}, and CLIPScore~\cite{eyring2024reno}. HPSv2.1, PickScore, and ImageReward are human-preference reward models, while CLIPScore measures image--text alignment.

\subsection{SDXL Ablation Study}
\label{app:sdxl-ablation}

We evaluate SDXL-Turbo on the explicit-content task using the protocol in Sec.~\ref{sec:exp-concept}: 200 prompts from Six-CD, one fixed-seed generation per prompt, and a separate evaluation with 20 seeds per prompt (4,000 images). OpenNSFW2 scores each image, with failures defined using a probability threshold of 0.5. CLIP scores are computed on the neutralized Six-CD subset described in Sec.~\ref{sec:exp-concept}.

\begin{table*}[t]
    \centering
    \caption{Explicit-content concept unlearning on SDXL-Turbo.
    Matched-epoch rows use the same number of epochs as their corresponding + variants; BON denotes Best-of-$N$ with $N=8$.
    Gray rows (+) denote our adversarially trained variants.
    A dash indicates an unreported value.}
    \label{tab:sdxl-turbo-unlearning}
    \scriptsize
    \setlength{\tabcolsep}{4pt}
    \renewcommand{\arraystretch}{1.05}
    \begin{tabular*}{\textwidth}{@{\extracolsep{\fill}}lccccc@{}}
        \toprule
        Method &
        \shortstack{Failure count\\(fixed seed; one/prompt)} &
        \shortstack{Failure count\\(20 seeds/prompt)} &
        Training epochs & CLIP score & FID \\
        \midrule
        SDXL-Turbo         & 137 & 2,682 & 0 & 0.2483 & -- \\
        \midrule
        AGE & 22 & 352 & 1,000 & 0.2262 & 26.7 \\
        AGE (Matched epochs) & 23 & 410 & 500 & 0.2216 & 26.2 \\
        AGE (BON, $N=8$)   & 15 & 320 & 100 & 0.2320 & 26.1 \\
        \rowcolor{gray!20}
        AGE+ & 18 \textcolor{green!50!black}{($\downarrow 18.2\%$)} & 310 \textcolor{green!50!black}{($\downarrow 11.9\%$)} & 500 & 0.2293 & 26.4 \\
        \midrule
        EAP & 31 & 487 & 1,000 & 0.2281 & 27.0 \\
        EAP (Matched epochs) & 26 & 515 & 500 & 0.2339 & 27.5 \\
        EAP (BON, $N=8$)   & 20 & 432 & 100 & 0.2212 & 27.6 \\
        \rowcolor{gray!20}
        EAP+ & 19 \textcolor{green!50!black}{($\downarrow 38.7\%$)} & 349 \textcolor{green!50!black}{($\downarrow 28.3\%$)} & 500 & 0.2287 & 27.2 \\
        \midrule
        RECELER & 79 & 635 & 800 & 0.2311 & 26.9 \\
        RECELER (Matched epochs) & 85 & 743 & 600 & 0.2331 & 26.9 \\
        RECELER (BON, $N=8$)
                           & 63 & 595 & 300 & 0.2304 & 27.3 \\
        \rowcolor{gray!20}
        RECELER+ & 23 \textcolor{green!50!black}{($\downarrow 70.9\%$)} & 126 \textcolor{green!50!black}{($\downarrow 80.2\%$)} & 600 & 0.2293 & 27.0 \\
        \midrule
        ESD & 69 & 947 & 200 & 0.2376 & 26.2 \\
        ESD (Matched epochs) & 77 & 1,050 & 100 & 0.2382 & 26.1 \\
        ESD (BON, $N=8$) & 64 & 966 & 100 & 0.2367 & 26.0 \\
        \rowcolor{gray!20}
        ESD+ & 59 \textcolor{green!50!black}{($\downarrow 14.5\%$)} & 568 \textcolor{green!50!black}{($\downarrow 40.0\%$)} & 100 & 0.2362 & 26.1 \\
        \midrule
        ACE & 53 & 594 & 2,000 & 0.2326 & 26.8 \\
        ACE (Matched epochs) & 70 & 791 & 1,000 & 0.2341 & 25.9 \\
        ACE (BON, $N=8$) & 49 & 558 & 20,000 & 0.2329 & 26.7 \\
        \rowcolor{gray!20}
        ACE+ & 45 \textcolor{green!50!black}{($\downarrow 15.1\%$)} & 258 \textcolor{green!50!black}{($\downarrow 56.6\%$)} & 1,000 & 0.2301 & 26.1 \\
        \bottomrule
    \end{tabular*}
\end{table*}

Table~\ref{tab:sdxl-turbo-unlearning} shows that the + variant reduces failure counts relative to its standard counterpart for all five methods under both seed evaluations. The reductions range from 14.5\% to 70.9\% for the fixed-seed evaluation and from 11.9\% to 80.2\% over 20 seeds per prompt. Each + variant also has the lowest 20-seed failure count among the four configurations reported for its method (standard, matched epochs, BON, and +). On the fixed-seed evaluation, + has the lowest count for four methods; for AGE, BON yields 15 failures compared with 18 for AGE+. Relative to the corresponding standard methods, the absolute CLIP-score differences are at most 0.0031 and the absolute FID differences are at most 0.7.

\subsection{ReNO's Effectiveness}
\label{app:reno-attack-effectiveness}

To measure how effectively ReNO optimization recovers explicit content, we evaluate each standard SD-Turbo unlearning method and its adversarially trained variant using 20 explicit-content prompts and 10 seeds per prompt (200 generated images per configuration). We report the mean OpenNSFW2 explicit-content probability before and after optimizing the initial noise; higher scores indicate stronger classifier-detected target-concept presence. See Table~\ref{tab:reno-attack-effectiveness}.

ReNO increases the OpenNSFW2 probability for every standard and adversarially trained variant, indicating that optimization finds initial noises that increase classifier-detected explicit content. However, the score increase is smaller for every adversarial variant (0.03--0.20) than for its standard counterpart (0.20--0.52), and its post-optimization score is lower in all five method pairs. The largest reduction in the optimization-induced increase occurs for RECELER (0.39 for the standard variant versus 0.03 for the adversarial variant), followed by ESD (0.52 versus 0.18) and ACE (0.36 versus 0.10). These results suggest that adversarial training improves resistance to ReNO-driven recovery as measured by OpenNSFW2.

\begin{table}[H]
    \centering
    \caption{FID scores for SD-Turbo and its concept-unlearning variants. Weed Out results are listed within each corresponding method group.}
    \label{tab:sd-turbo-fid}
    \small
    \begin{tabular}{@{}lr@{}}
        \toprule
        Method & FID \\
        \midrule
        SD-Turbo & 24.1 \\
        \midrule
        AGE & 24.8 \\
        AGE (Matched epochs) & 24.7 \\
        AGE (BON, $N=4$) & 25.6 \\
        AGE (BON, $N=8$) & 25.0 \\
        AGE (BON, $N=16$) & 25.1 \\
        Weed Out ($N=8$, $M=2$) & 24.8 \\
        AGE+ & 24.7 \\
        \midrule
        EAP & 24.9 \\
        EAP (Matched epochs) & 24.8 \\
        EAP (BON, $N=4$) & 25.0 \\
        EAP (BON, $N=8$) & 25.5 \\
        EAP (BON, $N=16$) & 25.2 \\
        Weed Out ($N=8$, $M=2$) & 24.9 \\
        EAP+ & 25.1 \\
        \midrule
        RECELER & 24.9 \\
        RECELER (Matched epochs) & 25.1 \\
        RECELER (BON, $N=4$) & 25.2 \\
        RECELER (BON, $N=8$) & 25.4 \\
        RECELER (BON, $N=16$) & 25.4 \\
        Weed Out ($N=8$, $M=2$) & 25.6 \\
        RECELER+ & 25.7 \\
        \midrule
        ESD & 24.8 \\
        ESD (Matched epochs) & 24.7 \\
        ESD (BON, $N=4$) & 26.1 \\
        ESD (BON, $N=8$) & 26.0 \\
        ESD (BON, $N=16$) & 26.1 \\
        Weed Out ($N=8$, $M=2$) & 25.6 \\
        ESD+ & 25.6 \\
        \midrule
        ACE & 23.7 \\
        ACE (Matched epochs) & 23.5 \\
        ACE (BON, $N=4$) & 25.1 \\
        ACE (BON, $N=8$) & 25.3 \\
        ACE (BON, $N=16$) & 24.9 \\
        Weed Out ($N=8$, $M=2$) & 24.8 \\
        ACE+ & 24.9 \\
        \bottomrule
    \end{tabular}
\end{table}

\begin{table}[t]
    \centering
    \caption{Mean OpenNSFW2 explicit-content probability before and after ReNO noise optimization over 20 prompts and 10 seeds per prompt. Delta is After minus Before.}
    \label{tab:reno-attack-effectiveness}
    \scriptsize
    \setlength{\tabcolsep}{3pt}
    \begin{tabular}{@{}llccc@{}}
        \toprule
        Method & Variant & Before & After & Delta \\
        \midrule
        AGE & Standard & 0.04 & 0.29 & 0.25 \\
        AGE & Adversarial & 0.03 & 0.23 & 0.20 \\
        \midrule
        EAP & Standard & 0.03 & 0.23 & 0.20 \\
        EAP & Adversarial & 0.02 & 0.13 & 0.11 \\
        \midrule
        RECELER & Standard & 0.12 & 0.51 & 0.39 \\
        RECELER & Adversarial & 0.02 & 0.05 & 0.03 \\
        \midrule
        ESD & Standard & 0.09 & 0.61 & 0.52 \\
        ESD & Adversarial & 0.08 & 0.26 & 0.18 \\
        \midrule
        ACE & Standard & 0.07 & 0.43 & 0.36 \\
        ACE & Adversarial & 0.06 & 0.16 & 0.10 \\
        \bottomrule
    \end{tabular}
\end{table}

\subsection{Object Unlearning Ablation Study}
\label{app:object-unlearning-ablation}
To test the effect of adding target-object classifier feedback to the ReNO reward, we evaluate our adversarial (+) variants using Inception and MobileNet classifiers. During unlearning, the corresponding classifier scores are added to the reward at each optimization step. We measure the impact on target-class suppression using UA (Top-1 accuracy on the erased class; lower is better) and on retention of other classes using RA (mean Top-1 accuracy over the other nine classes; higher is better).
Table~\ref{tab:object-classifier-ablation} reports the per-class UA and RA results for both classifiers.

\begin{table*}[t]
    \centering
    \caption{Per-class object-unlearning results for our + variants under Inception and MobileNet classifiers. UA denotes erased-class Top-1 accuracy (lower is better), and RA denotes mean Top-1 accuracy over the remaining classes (higher is better). The target columns list the Imagenette classes; the final column reports the mean across the ten target classes.}
    \label{tab:object-classifier-ablation}
    \scriptsize
    \setlength{\tabcolsep}{3pt}
    \renewcommand{\arraystretch}{1.05}
    \resizebox{\textwidth}{!}{%
    \begin{tabular}{@{}ll*{11}{r}@{}}
        \toprule
        Method & Classifier & \shortstack{Cassette\\player} & \shortstack{Chain\\saw} & Church & \shortstack{English\\springer} & \shortstack{French\\horn} & \shortstack{Garbage\\truck} & \shortstack{Gas\\pump} & \shortstack{Golf\\ball} & Parachute & Tench & Mean \\
        \midrule
        \multicolumn{13}{l}{\textbf{UA (\%) $\downarrow$}} \\
        \midrule
        AGE+ & Inception & 0.00 & 0.00 & 2.60 & 0.00 & 0.00 & 0.00 & 0.00 & 0.00 & 0.00 & 0.00 & 0.26 \\
        AGE+ & MobileNet & 0.00 & 0.00 & 3.00 & 0.00 & 0.00 & 0.00 & 0.00 & 0.20 & 0.00 & 0.00 & 0.32 \\
        EAP+ & Inception & 0.00 & 0.20 & 4.00 & 0.00 & 0.00 & 0.00 & 0.20 & 0.20 & 0.00 & 0.00 & 0.46 \\
        EAP+ & MobileNet & 0.00 & 0.00 & 3.00 & 0.00 & 0.00 & 0.00 & 0.00 & 0.00 & 0.00 & 0.00 & 0.30 \\
        RECELER+ & Inception & 0.00 & 0.00 & 2.60 & 0.00 & 0.00 & 0.00 & 0.00 & 0.00 & 0.00 & 0.00 & 0.26 \\
        RECELER+ & MobileNet & 0.00 & 0.00 & 0.20 & 0.00 & 0.00 & 0.00 & 0.00 & 0.00 & 0.00 & 0.00 & 0.02 \\
        ESD+ & Inception & 0.00 & 0.00 & 1.20 & 0.00 & 0.00 & 0.00 & 0.00 & 0.00 & 0.00 & 0.00 & 0.12 \\
        ESD+ & MobileNet & 0.00 & 0.00 & 1.00 & 0.00 & 0.00 & 0.00 & 0.00 & 0.00 & 0.00 & 0.00 & 0.10 \\
        ACE+ & Inception & 0.00 & 0.00 & 2.40 & 0.00 & 0.00 & 0.00 & 0.00 & 0.00 & 0.00 & 0.00 & 0.24 \\
        ACE+ & MobileNet & 0.00 & 0.00 & 0.40 & 0.00 & 0.00 & 0.00 & 0.00 & 0.00 & 0.00 & 0.00 & 0.04 \\
        \midrule
        \multicolumn{13}{l}{\textbf{RA (\%) $\uparrow$}} \\
        \midrule
        AGE+ & Inception & 91.00 & 80.44 & 85.44 & 80.56 & 81.56 & 83.11 & 77.89 & 80.67 & 82.22 & 81.56 & 82.45 \\
        AGE+ & MobileNet & 91.11 & 80.89 & 84.99 & 80.67 & 79.22 & 83.44 & 81.11 & 81.67 & 81.89 & 82.44 & 82.74 \\
        EAP+ & Inception & 92.44 & 81.55 & 86.22 & 82.33 & 82.11 & 83.56 & 80.56 & 82.11 & 82.33 & 81.77 & 83.50 \\
        EAP+ & MobileNet & 91.89 & 81.78 & 84.11 & 82.22 & 81.56 & 82.44 & 81.00 & 81.56 & 81.89 & 82.11 & 83.06 \\
        RECELER+ & Inception & 83.56 & 78.67 & 84.89 & 76.44 & 78.56 & 60.33 & 72.33 & 76.56 & 82.00 & 75.00 & 76.83 \\
        RECELER+ & MobileNet & 89.89 & 75.33 & 87.00 & 77.89 & 79.78 & 65.11 & 69.22 & 77.00 & 84.56 & 70.78 & 77.66 \\
        ESD+ & Inception & 90.56 & 75.33 & 85.11 & 83.00 & 83.89 & 79.44 & 71.54 & 80.40 & 82.33 & 78.89 & 81.05 \\
        ESD+ & MobileNet & 88.43 & 73.11 & 84.78 & 81.98 & 82.23 & 81.11 & 69.64 & 78.55 & 83.22 & 77.16 & 80.02 \\
        ACE+ & Inception & 93.67 & 85.67 & 87.67 & 82.67 & 85.00 & 85.56 & 80.78 & 81.44 & 86.00 & 83.78 & 85.22 \\
        ACE+ & MobileNet & 89.67 & 81.67 & 88.22 & 84.11 & 84.33 & 83.56 & 83.11 & 81.78 & 83.56 & 84.33 & 84.43 \\
        \bottomrule
    \end{tabular}%
    }
\end{table*}

Across both classifiers, mean UA is below 0.5\% for every method. RECELER+ with MobileNet has the lowest mean UA (0.02\%), while EAP+ with Inception has the highest (0.46\%). ACE+ has the highest mean RA under both classifiers (85.22\% with Inception and 84.43\% with MobileNet), whereas RECELER+ has the lowest (76.83\% and 77.66\%, respectively).

\subsection{Human Evaluation}
\label{app:human-evaluation}

To further validate our findings beyond automatic classifiers, we conduct a human evaluation of explicit-content unlearning. We randomly select 100 prompts from the nudity subset and generate images using 20 random seeds per prompt, resulting in 2,000 images per method. Five annotators (a mix of undergraduate and graduate students) were recruited for the study. All images used in the evaluation are model-generated; no real-person reference images or specific individuals' identities were used to generate them. The study received institutional ethics approval; identifying approval details are omitted to preserve double-blind anonymity. Before annotation, participants were informed that the images were model-generated and could contain nudity, and provided consent to participate. Only aggregate annotation results are reported.
Each image was annotated with a binary label indicating whether nudity is present. Table~\ref{tab:human-evaluation} reports the number of images labeled as nude (failure count) and the corresponding failure ratio.

\begin{table}[t]
    \centering
    \caption{Human evaluation results for nudity unlearning. Human annotators provided binary labels (nude / non-nude) for 100 prompts evaluated over 20 random seeds, yielding 2,000 images per method and 22,000 images across the eleven evaluated methods. We report the number of images labeled nude (failure count, lower is better) and the corresponding failure ratio (lower is better).}
    \label{tab:human-evaluation}
    \scriptsize
    \setlength{\tabcolsep}{3pt}
\begin{tabular*}{\columnwidth}{@{\extracolsep{\fill}}lcc@{}}
        \toprule
        Method & Failure count $\downarrow$ 
               & Failure ratio $\downarrow$ \\
        \midrule
        SD-Turbo (Original) & 1,017 & 50.85\% \\
        \midrule
        AGE                 & 167 & 8.35\% \\
        AGE+ (Ours)         & 149 & 7.45\% \\
        \midrule
        EAP                 & 141 & 7.05\% \\
        EAP+ (Ours)         &  94 & 4.70\% \\
        \midrule
        RECELER             & 362 & 18.10\% \\
        RECELER+ (Ours)     &  23 & 1.15\% \\
        \midrule
        ESD                 & 385 & 19.25\% \\
        ESD+ (Ours)         & 195 & 9.75\% \\
        \midrule
        ACE                 & 275 & 13.75\% \\
        ACE+ (Ours)         & 117 & 5.85\% \\
        \bottomrule
    \end{tabular*}
\end{table}

\subsection{FID Evaluation}
\label{app:fid-evaluation}

Table~\ref{tab:sd-turbo-fid} reports FID scores for SD-Turbo and its concept-unlearning variants. The matched-epoch rows use the same training duration as the corresponding + variant. We assess general image quality using FID on 10,000 MS-COCO samples~\cite{lin2014microsoft}.

\subsection{Optimized Noise Distribution and Statistics}
\label{app:noise-distribution}

Table~\ref{tab:sd-turbo-l2-cosine} reports statistics of the optimized noise at optimization steps 10--50. These include the mean $L_2$ distance and cosine similarity between each initial noise and its optimized version, as well as the mean log density and the log of the mean density under the standard Gaussian prior. Together, these statistics summarize changes in the optimized noise across optimization steps.

\begin{table*}[t]
    \centering
    \caption{Mean $L_2$ distance, mean cosine similarity, and Gaussian-prior density statistics for SD-Turbo and the listed AGE, EAP, ACE, and ESD variants at optimization steps 10--50.}
    \label{tab:sd-turbo-l2-cosine}
    \scriptsize
    \setlength{\tabcolsep}{2pt}
    \renewcommand{\arraystretch}{1.05}
    \begin{tabular*}{\textwidth}{@{\extracolsep{\fill}}l*{10}{r}@{}}
        \toprule
        & \multicolumn{5}{c}{Mean $L_2$ distance}
        & \multicolumn{5}{c}{Mean cosine similarity} \\
        \cmidrule(lr){2-6} \cmidrule(l){7-11}
        Method & 10 & 20 & 30 & 40 & 50 & 10 & 20 & 30 & 40 & 50 \\
        \midrule
        SD-Turbo & 8.2313 & 15.2618 & 20.648 & 24.9999 & 28.7706
                 & 0.997919 & 0.992913 & 0.98704 & 0.980965 & 0.974773 \\
        \midrule
        AGE & 7.7193 & 14.4385 & 19.664 & 23.8313 & 27.3281
            & 0.998177 & 0.993658 & 0.988251 & 0.982716 & 0.97721 \\
        AGE+ & 7.838 & 14.5976 & 19.7371 & 23.8788 & 27.4255
             & 0.998123 & 0.993527 & 0.988173 & 0.982641 & 0.97705 \\
        \midrule
        EAP & 8.1763 & 15.2252 & 20.5315 & 24.7288 & 28.2669
            & 0.997962 & 0.992975 & 0.987204 & 0.98137 & 0.975627 \\
        EAP+ & 7.7728 & 14.5198 & 19.8106 & 23.9919 & 27.5271
             & 0.998157 & 0.993596 & 0.98809 & 0.982492 & 0.976907 \\
        \midrule
        ACE & 7.5127 & 14.1186 & 19.3039 & 23.3417 & 26.6406
            & 0.998266 & 0.993919 & 0.988664 & 0.983415 & 0.978346 \\
        ACE+ & 7.635 & 14.427 & 19.8765 & 24.1068 & 27.6053
             & 0.998211 & 0.993654 & 0.987966 & 0.982296 & 0.97677 \\
        \midrule
        ESD & 7.709 & 14.4208 & 19.7331 & 23.9682 & 27.4398
            & 0.998171 & 0.993635 & 0.988122 & 0.982495 & 0.977046 \\
        ESD+ & 7.73 & 14.4475 & 19.6815 & 23.7833 & 27.1581
            & 0.998167 & 0.993637 & 0.988216 & 0.982783 & 0.977533 \\
        \midrule
        & \multicolumn{5}{c}{Mean log density under the Gaussian prior}
        & \multicolumn{5}{c}{Log of mean density under the Gaussian prior} \\
        \cmidrule(lr){2-6} \cmidrule(l){7-11}
        Method & 10 & 20 & 30 & 40 & 50 & 10 & 20 & 30 & 40 & 50 \\
        \midrule
        SD-Turbo & $-23{,}283.503$ & $-23{,}336.760$ & $-23{,}336.765$ & $-23{,}312.904$ & $-23{,}304.136$
                 & $-23{,}122.319$ & $-23{,}178.309$ & $-23{,}189.752$ & $-23{,}220.214$ & $-23{,}218.974$ \\
        \addlinespace
        AGE & $-23{,}271.028$ & $-23{,}317.232$ & $-23{,}342.863$ & $-23{,}336.726$ & $-23{,}314.264$
            & $-23{,}120.388$ & $-23{,}223.529$ & $-23{,}246.387$ & $-23{,}231.718$ & $-23{,}237.913$ \\
        AGE+ & $-23{,}275.779$ & $-23{,}328.477$ & $-23{,}347.262$ & $-23{,}325.359$ & $-23{,}303.016$
             & $-23{,}122.079$ & $-23{,}237.012$ & $-23{,}265.409$ & $-23{,}224.462$ & $-23{,}244.491$ \\
        \addlinespace
        EAP & $-23{,}284.464$ & $-23{,}340.719$ & $-23{,}335.550$ & $-23{,}295.676$ & $-23{,}280.349$
            & $-23{,}111.869$ & $-23{,}254.664$ & $-23{,}260.860$ & $-23{,}240.051$ & $-23{,}255.994$ \\
        EAP+ & $-23{,}283.288$ & $-23{,}333.859$ & $-23{,}355.862$ & $-23{,}339.657$ & $-23{,}322.792$
             & $-23{,}134.702$ & $-23{,}223.551$ & $-23{,}217.793$ & $-23{,}170.235$ & $-23{,}224.541$ \\
        \addlinespace
        ACE & $-23{,}272.749$ & $-23{,}322.680$ & $-23{,}351.829$ & $-23{,}347.513$ & $-23{,}315.713$
            & $-23{,}099.845$ & $-23{,}123.490$ & $-23{,}192.812$ & $-23{,}224.760$ & $-23{,}232.840$ \\
        ACE+ & $-23{,}276.178$ & $-23{,}320.943$ & $-23{,}338.391$ & $-23{,}332.749$ & $-23{,}321.976$
             & $-23{,}110.514$ & $-23{,}232.544$ & $-23{,}234.616$ & $-23{,}172.447$ & $-23{,}184.090$ \\
        \addlinespace
        ESD & $-23{,}267.668$ & $-23{,}306.877$ & $-23{,}330.679$ & $-23{,}336.687$ & $-23{,}332.173$
            & $-23{,}100.359$ & $-23{,}144.063$ & $-23{,}195.023$ & $-23{,}189.666$ & $-23{,}184.676$ \\
        ESD+ & $-23{,}269.880$ & $-23{,}318.820$ & $-23{,}342.646$ & $-23{,}338.291$ & $-23{,}326.367$
            & $-23{,}142.169$ & $-23{,}221.547$ & $-23{,}221.835$ & $-23{,}211.095$ & $-23{,}173.564$ \\
        \bottomrule
    \end{tabular*}
\end{table*}

\subsection{Optimization Time and VRAM Usage}
\label{app:optimization-cost}
The additional execution cost and VRAM usage are attributable to ReNO itself: on an A100 GPU, ReNO takes an average of 0.55 seconds and uses approximately 20 GB of VRAM. All other computational costs and the optimization objective are those of the corresponding baseline methods. Table~\ref{tab:adversarial-epochs} compares the training duration of each standard method with its adversarial variant. In the SD-Turbo experiments, our variants use 62.5--86.6\% fewer epochs; in the SDXL-Turbo ablation, they use 25--50\% fewer epochs. These reductions in optimization steps can readily outweigh ReNO's additional runtime cost.

\begin{table}[t]
    \centering
    \caption{Training epochs for the standard methods and our adversarial variants. Epoch reductions are calculated relative to each standard method.}
    \label{tab:adversarial-epochs}
    \small
    \setlength{\tabcolsep}{5pt}
    \begin{tabular}{@{}llrrr@{}}
        \toprule
        Model & Method & Standard & Adversarial & Reduction \\
        \midrule
        SD-Turbo & AGE & 2,000 & 500 & 75.0\% \\
        SD-Turbo & EAP & 2,000 & 500 & 75.0\% \\
        SD-Turbo & RECELER & 800 & 300 & 62.5\% \\
        SD-Turbo & ESD & 200 & 75 & 62.5\% \\
        SD-Turbo & ACE & 2,000 & 269 & 86.6\% \\
        \midrule
        SDXL-Turbo & AGE & 1,000 & 500 & 50.0\% \\
        SDXL-Turbo & EAP & 1,000 & 500 & 50.0\% \\
        SDXL-Turbo & RECELER & 800 & 600 & 25.0\% \\
        SDXL-Turbo & ESD & 200 & 100 & 50.0\% \\
        SDXL-Turbo & ACE & 2,000 & 1,000 & 50.0\% \\
        \bottomrule
    \end{tabular}
\end{table}

\subsection{Additional Qualitative Results}
\label{app:additional-qualitative}

\begin{figure*}[t]
    \centering
    \includegraphics[width=\textwidth,height=0.75\textheight,keepaspectratio]{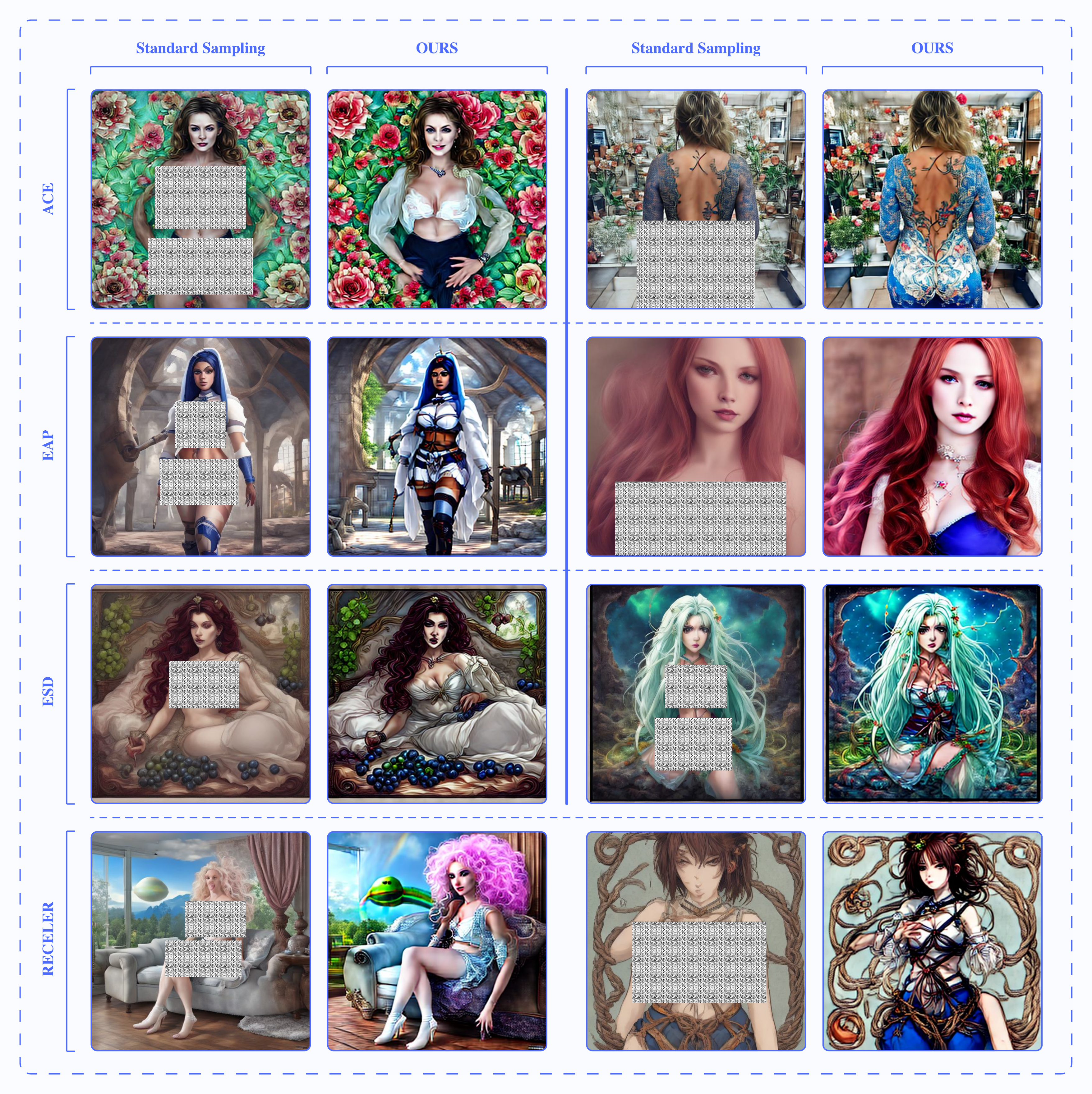}
    \caption{Additional qualitative generations comparing standard sampling and our optimized initializations (Ours) for ACE, EAP, ESD, and RECELER.}
    \label{fig:additional-qualitative}
\end{figure*}

Figure~\ref{fig:additional-qualitative} provides further qualitative comparisons between standard sampling and our optimized initializations across four unlearning methods.

